\documentclass[pre,aps,twocolumn]{revtex4-2}

\usepackage{amsmath}
\usepackage{amssymb}
\usepackage{graphicx}
\usepackage{dcolumn}
\usepackage{bm}
\usepackage{soul}
\usepackage{xcolor}
\usepackage{hyperref}

\def\eqq#1{Eq.~(\ref{#1})}
\def\eq#1{(\ref{#1})}

\def\f#1{Fig.~\ref{#1}}
\def\ff#1{Figs.~\ref{#1}}

\def\c#1{~\cite{#1}}

\begin{document}

\title{Optimizing thermodynamic trajectories using evolutionary and gradient-based reinforcement learning}

\author{Chris Beeler$^{1,2}$}
\email{christopher.beeler@uottawa.ca}
\author{Uladzimir Yahorau$^3$}
\author{Rory Coles$^{4}$}
\author{Kyle Mills$^{3,5}$}
\author{Stephen Whitelam$^6$}
\author{Isaac Tamblyn$^{1,3,5}$}
\email{isaac.tamblyn@uottawa.ca}
\affiliation{$^1$University of Ottawa, Ottawa, ON, Canada\\
$^2$National Research Council of Canada, Ottawa, ON, Canada\\
$^3$University of Ontario Institute of Technology, Oshawa, ON, Canada\\
$^4$University of Victoria, Victoria, BC, Canada\\
$^5$Vector Institute for Artificial Intelligence, Toronto, Ontario, Canada\\
$^6$Molecular Foundry, Lawrence Berkeley National Laboratory, Berkeley, CA, USA}

\date{November 22, 2021}

\widetext
\begin{abstract}
Using a model heat engine, we show that neural network-based reinforcement learning can identify thermodynamic trajectories of maximal efficiency. We consider both gradient and gradient-free reinforcement learning. We use an evolutionary learning algorithm to evolve a population of neural networks, subject to a directive to maximize the efficiency of a trajectory composed of a set of elementary thermodynamic processes; the resulting networks learn to carry out the maximally-efficient Carnot, Stirling, or Otto cycles. When given an additional irreversible process, this evolutionary scheme learns a previously unknown thermodynamic cycle. Gradient-based reinforcement learning is able to learn the Stirling cycle, whereas an evolutionary approach achieves the optimal Carnot cycle. Our results show how the reinforcement learning strategies developed for game playing can be applied to solve physical problems conditioned upon path-extensive order parameters.

\end{abstract}

\maketitle
\section{Introduction}
Games, whether played on a board, such as chess or Go, or played on the computer, are a major component of human culture\c{roberts1959games}. In the language of physics, instances of a game are {\em trajectories}, time-ordered sequences of elementary steps. The outcome of a game is a path-extensive order parameter determined by the entire history of the trajectory. Playing games was once the preserve of human beings, but machine learning methods, more specifically reinforcement learning, now outperform the most talented humans in all the aforementioned examples\c{QL,DQN,Atari,Atari2600,Actor,DeepMind,MuJoCo,MDP,Rogue,NFQ,Soccer,PPO,Guber,OpenAI,VizDoom,VizDoom2,Go,Go2}. While there are more scientific examples of reinforcement learning\c{mills2020finding, andreasson2019quantum, zhou2019optimization, radaideh2020physics, badloe2020biomimetic, popova2018deep}, we choose to compare with games as they are the easiest applications to understand. Motivated by the correspondence between games and trajectories, it is natural to ask how the machine-learning and optimal control methods that have mastered game-playing might be applied to understand physical processes whose outcomes are path-extensive quantities.

There are many examples of such processes. For instance, the success or failure of molecular self-assembly is determined by a time history of elementary dynamical processes, including the binding and unbinding of particles\c{de2015crystallization,hagan2006dynamic,wilber2007reversible,whitelam2015statistical}. Dynamical systems, such as chemical networks and molecular machines\c{gillespie2007stochastic,mcgrath2017biochemical,seifert2012stochastic,brown2017allocating}, are characterized by time-extensive observables, such as work or entropy production\c{seifert2005entropy,lecomte2007thermodynamic,ritort2008nonequilibrium,garrahan2009first,speck2012large,lecomte2010current,harris2015fluctuations}. In none of these cases do we possess a complete theoretical or practical understanding of how to build an arbitrary structure or maximize the efficiency of an arbitrary machine. Traditional methods of inquiry in physics focus on applying physical intuition and the manipulation and simulation of equations; perhaps machine learning can provide us with further insight into physical problems of a path-extensive nature.

\begin{figure*}
   \centering
   \includegraphics[width=\linewidth]{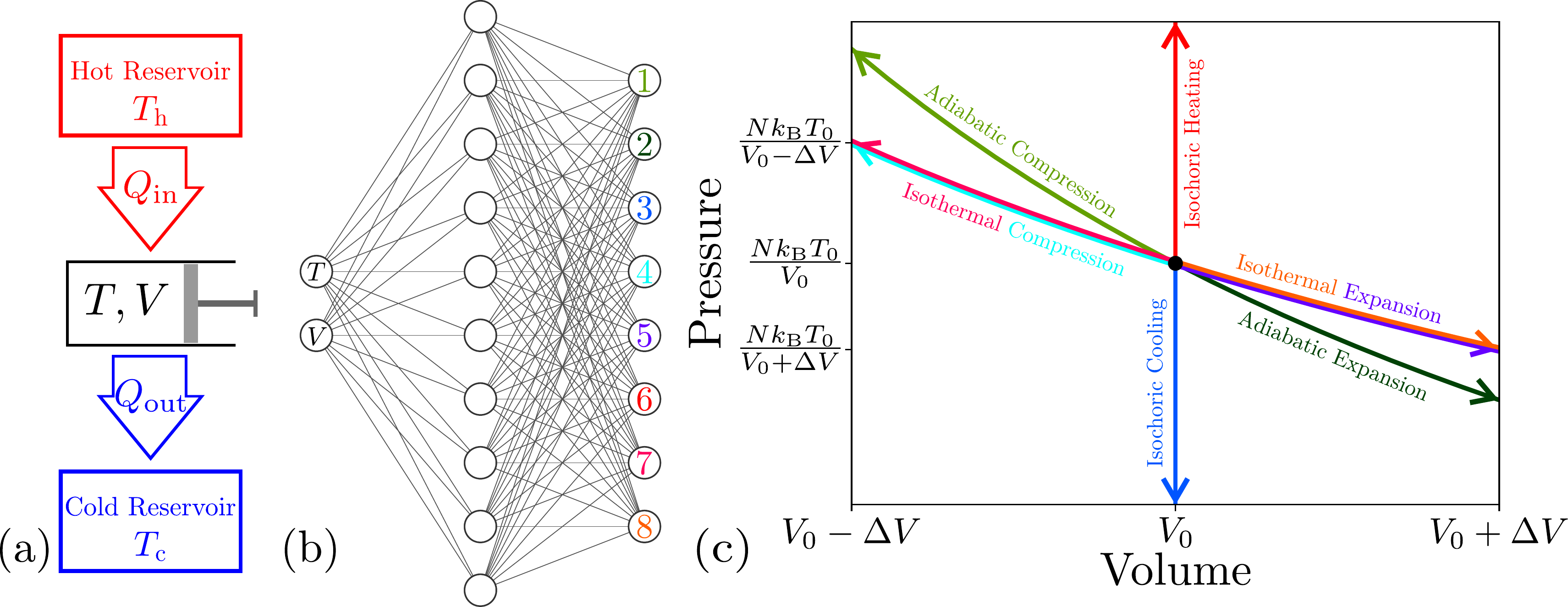}
   \caption{(a) Model heat engine and (b) the neural network that controls it. Note that the diagram of the neural network has two additional input nodes for the gradient-based reinforcement learning method. (c) A summary of the actions in $P$-$V$ space available to the network; see Tab. \ref{tab:action}.}
   \label{fig1}
\end{figure*}

Motivated by this speculation, we show here that neural network-based reinforcement learning can maximize the efficiency of the simplest type of physical trajectories, the deterministic, quasi-static ones of classical thermodynamics. We introduce a model heat engine characterized by a set of thermodynamic state variables. A neural network takes as input the current observation of the engine and chooses one of a set of basic thermodynamic processes to produce a new observation; this change comprises one step of a trajectory. We chose this simple and well-known system for pedagogical purposes. Using an easy-to-understand system allows us to focus on how reinforcement learning methods can be applied to physics problems. We generate a set of trajectories of fixed length using a set of networks whose parameters are initially randomly chosen and optimized using two different methods. In the first method (gradient-free), we retain and mutate only those networks whose trajectories show the greatest thermal efficiency. Repeating this evolutionary process many times results in networks whose trajectories reproduce the maximally efficient Carnot, Stirling, or Otto cycles, depending upon which basic thermodynamic processes are allowed. This evolutionary procedure can also learn previously unknown thermodynamic cycles if new processes are allowed. In the second method (gradient-based), we update the network parameters using gradient-based reinforcement learning. In this study, we explore how a reinforcement learning problem must be framed and the types of solutions acquired by these two methods. We also compare the advantages and disadvantages of each in finding a solution to this problem.

\section{Model heat engine and thermodynamic trajectories}
In \f{fig1}(a) we show a model heat engine, a device able to transform thermal energy into work\c{Callen_1985, Carnot}. The engine consists of a working substance, which we assume to be a monatomic ideal gas, housed within a frictionless container of variable volume $V$, whose minimum and maximum values are $V_{\rm min}$ and $V_{\rm max}$, respectively. The working substance may be connected to a hot or cold reservoir held at temperature, $T_{\rm h}=500$ K and $T_{\rm c}=300$ K, respectively. For the gradient-free method, the instantaneous observation $s_{t}$ of the system at time $t$ is then specified by the volume-temperature vector $s_{t}=(V,T)$, with the pressure of the system fixed by the ideal-gas equation $PV = N k_{\rm B}T$\c{Ideal}. Further on, we will discuss the reasons this formulation of the problem cannot be used for gradient-based reinforcement learning methods.

To evolve the heat engine we use the neural network shown in \f{fig1}(b). The network is a nonlinear function that takes as input the current observation $s_{t}$ of the system, and outputs the probabilities $\pi_{\theta}(a_{t} | s_{t})$ of moving to any new observation $s_{t+1}$ through a thermodynamic process $a_{t} \in \{a_{1}, a_{2}, \dots, a_{M}\}$ (in the language of reinforcement learning this mapping is called a {\em policy}\c{RL}). The symbol $\theta$ denotes the internal parameters of the network, discussed shortly. Here we consider deterministic evolution through configuration space, with $\pi_{\theta}(a_{t}^{*} | s_{t})$ equal to 1 for a chosen process $a_{t}^{*}$, and equal to zero otherwise. Enacting the chosen process corresponds to one step of a trajectory. Given an initial observation $s_0$, $K$ applications of the network produces a trajectory $\omega = s_0 \to s_1 \to \cdots \to s_K$  of $K$ steps through configuration space. We focus on trajectories of fixed length ($K=200$).

The elementary actions available to the network correspond to the basic thermodynamic processes shown in Tab. \ref{tab:action}, summarized graphically in \f{fig1}(c). These processes include reversible compression and expansion, along isotherms or adiabats, and reversible temperature changes along isochores. Implicitly this means infinitely many heat baths spaced between $T_{\rm h}$ and $T_{\rm c}$ are available, a condition, which we shall prove in Appendix \ref{sec:proof}, does not undermine Carnot's Theorem of maximum efficiency. All compression and expansion processes are performed using a fixed change in volume. If an isothermal process at $T_{\rm h}$ (or $T_{\rm c}$) is selected when the system is not at the correct temperature, then isochoric heating (or cooling) is performed first to reach the necessary temperature for the isothermal process to occur. Note that the isochoric processes are not used in the Carnot cycle, but are required to make approximations of it when using fixed changes in volume because the system must reach specific volumes in order to be adiabatically heated from $T_{\rm c}$ to $T_{\rm h}$ and cooled from $T_{\rm h}$ to $T_{\rm c}$.

Upon undertaking any action $s_{t} \to s_{t+1}$, we record the resulting changes of work, $\Delta W_{s_{t}s_{t+1}}$, and heat input from the hot reservoir, $\Delta Q^{\rm in}_{s_{t} s_{t+1}} = \Delta Q_{s_{t}s_{t+1}} H(\Delta Q_{s_{t}s_{t+1}} \delta_{T_{\rm f}, T_{\rm h}})$; these are listed in Tab. \ref{tab:action}. Here $H(\cdot)$ is the Heaviside function, equal to 1 for positive values of $\Delta Q_{s_{t}s_{t+1}}$ and 0 otherwise, and $T_{\rm f}$ is the temperature of the system following the move. $\delta_{\alpha, \beta}$ is the Kronecker delta (1 if $\alpha=\beta$ and 0 otherwise). We define the thermodynamic efficiency of a $K$-step trajectory as
\begin{equation} \label{eq:eff_calc}
\eta_K \equiv \frac{\sum_{t=0}^{K-1} \Delta W_{s_t s_{t+1}}}{\sum_{t=0}^{K-1} \Delta Q^{\rm in}_{s_{t} s_{t+1}}}.
\end{equation}
The thermal efficiency, a path-extensive quantity, is used as a means of ranking trajectories, and the networks that generate them, during our evolutionary learning procedure. The neural network selects processes deterministically based on observations, therefore once a trajectory produces a single cycle, that cycle will repeat until the maximum number of trajectory steps have occurred. For this reason, the maximum value $\eta = \max_K \eta_K$ for all $K$ points along a long trajectory is sufficient to identify efficient thermodynamic cycles. We could terminate a trajectory after it produces a single cycle, however we chose this way for consistency with the gradient-based reinforcement learning method used where we do require multiple cycles.
\begin{table*}
\caption{All possible actions that can be taken on our model heat engine and their corresponding $\Delta W$ and $\Delta Q$ equations.}
\label{tab:action}
\begin{tabular}{|c|c|c|} \hline
Action & $\Delta W$ & $\Delta Q$\\ \hline
Adiabatic Compression & $-\frac{3}{2}Nk_{\rm B}T_{\rm i}\left(\left(\frac{V_{\rm i}}{V_{\rm f}}\right)^{\frac{2}{3}}-1\right)$ & 0\\ \hline
Adiabatic Expansion & $-\frac{3}{2}Nk_{\rm B}T_{\rm i}\left(\left(\frac{V_{\rm i}}{V_{\rm f}}\right)^{\frac{2}{3}}-1\right)$ & 0\\ \hline
Isothermal Compression at $T_{\rm h}$ ($T=T_{\rm h}$) & $Nk_{\rm B}T_{\rm h}\log\left(\frac{V_{\rm f}}{V_{\rm i}}\right)$ & $Nk_{\rm B}T_{\rm h}\log\left(\frac{V_{\rm f}}{V_{\rm i}}\right)$\\ \hline
Isothermal Expansion at $T_{\rm h}$ ($T=T_{\rm h}$) & $Nk_{\rm B}T_{\rm h}\log\left(\frac{V_{\rm f}}{V_{\rm i}}\right)$ & $Nk_{\rm B}T_{\rm h}\log\left(\frac{V_{\rm f}}{V_{\rm i}}\right)$\\ \hline
Isothermal Compression at $T_{\rm h}$ ($T{\neq}T_{\rm h}$) & $Nk_{\rm B}T_{\rm h}\log\left(\frac{V_{\rm f}}{V_{\rm i}}\right)$ & $Nk_{\rm B}T_{\rm h}\log\left(\frac{V_{\rm f}}{V_{\rm i}}\right) + \frac{3}{2}Nk_{\rm B}\left(T_{\rm h} - T_{\rm i}\right)$\\ \hline
Isothermal Expansion at $T_{\rm h}$ ($T{\neq}T_{\rm h}$) & $Nk_{\rm B}T_{\rm h}\log\left(\frac{V_{\rm f}}{V_{\rm i}}\right)$ & $Nk_{\rm B}T_{\rm h}\log\left(\frac{V_{\rm f}}{V_{\rm i}}\right) + \frac{3}{2}Nk_{\rm B}\left(T_{\rm h} - T_{\rm i}\right)$\\ \hline
Isothermal Compression at $T_{\rm c}$ ($T=T_{\rm c}$) & $Nk_{\rm B}T_{\rm c}\log\left(\frac{V_{\rm f}}{V_{\rm i}}\right)$ & $Nk_{\rm B}T_{\rm c}\log\left(\frac{V_{\rm f}}{V_{\rm i}}\right)$\\ \hline
Isothermal Expansion at $T_{\rm c}$ ($T=T_{\rm c}$) & $Nk_{\rm B}T_{\rm c}\log\left(\frac{V_{\rm f}}{V_{\rm i}}\right)$ & $Nk_{\rm B}T_{\rm c}\log\left(\frac{V_{\rm f}}{V_{\rm i}}\right)$\\ \hline
Isothermal Compression at $T_{\rm c}$ ($T{\neq}T_{\rm c}$) & $Nk_{\rm B}T_{\rm c}\log\left(\frac{V_{\rm f}}{V_{\rm i}}\right)$ & $Nk_{\rm B}T_{\rm c}\log\left(\frac{V_{\rm f}}{V_{\rm i}}\right) + \frac{3}{2}Nk_{\rm B}\left(T_{\rm c} - T_{\rm i}\right)$\\ \hline
Isothermal Expansion at $T_{\rm c}$ ($T{\neq}T_{\rm c}$) & $Nk_{\rm B}T_{\rm c}\log\left(\frac{V_{\rm f}}{V_{\rm i}}\right)$ & $Nk_{\rm B}T_{\rm c}\log\left(\frac{V_{\rm f}}{V_{\rm i}}\right) + \frac{3}{2}Nk_{\rm B}\left(T_{\rm c} - T_{\rm i}\right)$\\ \hline
Isochoric Heating & 0 & $\frac{3}{2}Nk_{\rm B}\left(T_{\rm h} - T_{\rm i}\right)$\\ \hline
Isochoric Cooling & 0 & $\frac{3}{2}Nk_{\rm B}\left(T_{\rm c} - T_{\rm i}\right)$\\ \hline
\end{tabular}
\end{table*}

\section{Neural network-based policy}
The neural network, which contains two layers of tunable weights, performs computations as follows. Two input neurons receive the current observation $s_{t}$, and the output is comprised of $M \leq 8$ neurons, each corresponding to one of the actions shown in Tab. \ref{tab:action} (in some simulations we prohibit certain actions). The network possesses one hidden layer of $1024$ neurons, each connected to every input and output neuron. This architecture was chosen after several tests as it can produce optimal results while still keeping a low computational cost relative to larger (more parameters) networks that produce similar results. Let the indices $i \in \{1,2\}$, $j \in \{1,\dots,1024\}$, and $k \in \{1,\dots,M\}$ label the neurons of the input, hidden, and output layers, respectively. The input $I_i$ of the two nodes $i=0,1$ of the input layer are, respectively, scaled versions of the current temperature $(T-T_{\rm c})/(T_{\rm h}-T_{\rm c}) \in [0,1]$ and volume $(V-V_{\rm min})/(V_{\rm max}-V_{\rm min}) \in [0,1]$ of the system. We set the output signal $S_i$ of each input-layer node as $S_i=I_i$.

The input $I_j$ to neuron $j$ in the hidden layer is
\begin{equation}
I_j = \sum_{i=1}^2 S_i w_{ij},
\end{equation}
where the sum runs over the two neurons in the input layer, and $w_{ij}$ is the weight of the connection between nodes $i$ and $j$. We set the output signal $S_j$ of neuron $j$ to be
\begin{equation}
S_j = \frac{1}{2} \left[ \tanh\left( I_j+b_j \right) \right],
\end{equation}
where $b_j$ is a bias associated with neuron $j$. 

The input $I_k$ to neuron $k$ in the output layer is
\begin{equation}
I_k = \sum_{j=1}^{1024} S_j w_{jk},
\end{equation}
where the sum runs over all 1024 neurons of the hidden layer. Finally, we take the output signal $S_k$ from each output-layer neuron to be equal to $I_k$. To choose an action we pick the output neuron, $k^\star$, with the largest value of $S_k$. Given a current observation $s_{t}$, this action $a_{t}^{\star}$ defines a new observation $s_{t+1}^\star$ via Tab. \ref{tab:action}. The probability $\pi_{\theta}(a_{t}^{\star} | s_{t})$ is then unity, and all other $\pi_{\theta}(a_{t} | s_{t})$ are zero. We denote by $\theta = \{\{w\},\{b\}\}$ the set of all weights and biases of the network. Initially each weight and bias is chosen from a Gaussian distribution with zero mean and unit variance.

\begin{figure}
   \centering
   \includegraphics[width=\linewidth]{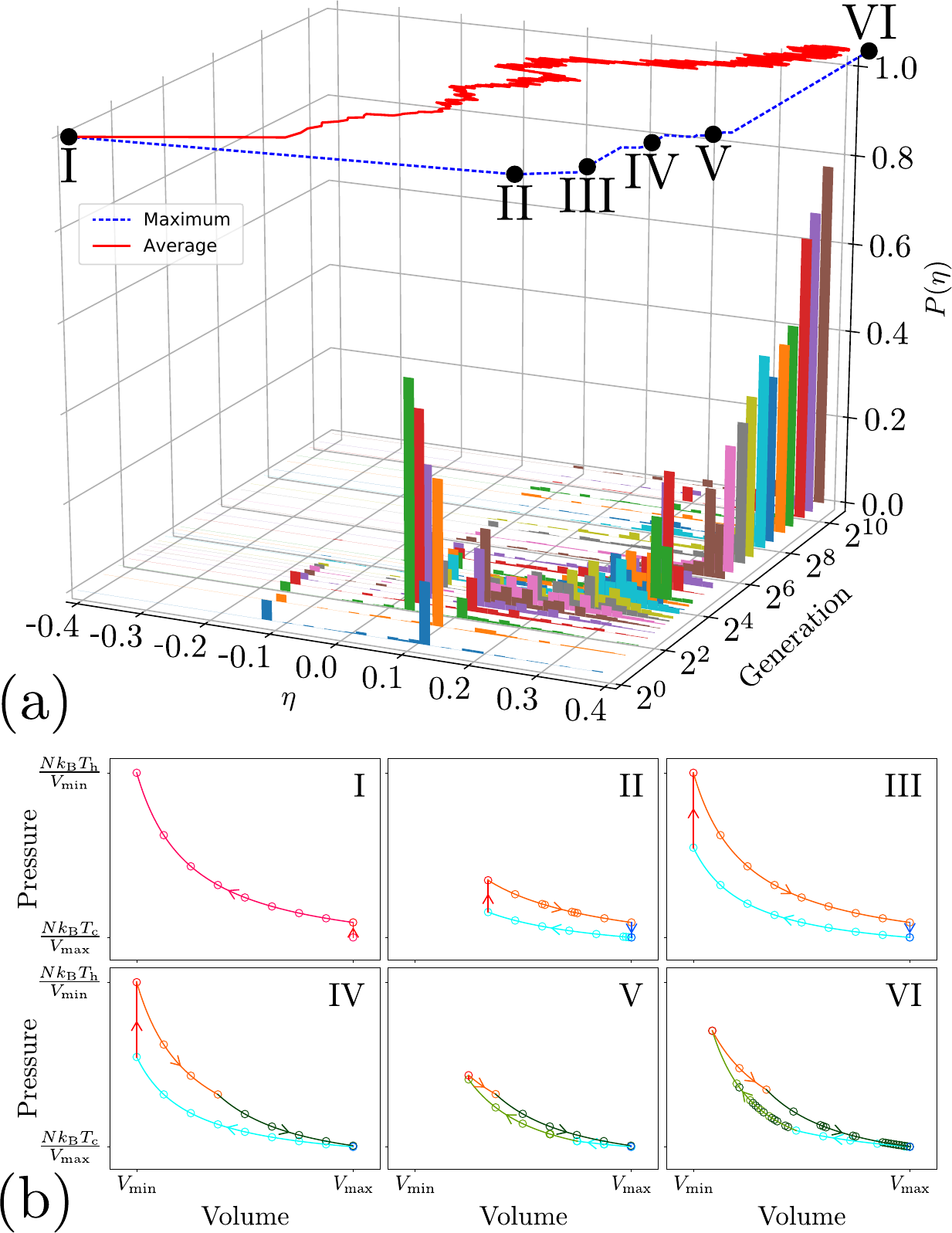} 
   \caption{(a) The evolution, as a function of generation number, of the probability distribution $P(\eta)$ of efficiencies $\eta$ of trajectories of the model heat engine. The maximum and average efficiency of the population are shown above. The Carnot efficiency is $\eta_{\rm max} = 0.4$. (b) Trajectories in $P$-$V$ space produced by the best-performing networks in generations $2^{0}$, $2^{1}$, $2^{2}$, $2^{4}$, $2^{5}$, and $2^{12}$, in the boxes labeled I-VI, respectively. The colors (shades), curvature, and direction of the branches correspond to the processes shown in \f{fig1}. Highly-evolved networks enact the Carnot cycle.}
   \label{fig2}
\end{figure}

\section{Evolutionary learning dynamics (gradient-free)}
With the thermodynamic system and means of evolving it defined, we introduce an evolutionary learning dynamics designed to produce networks able to propagate efficient thermodynamic trajectories. We start with a population of 100 networks, with the internal parameters $\theta$ of each initialized in the random fashion described above. We name this population generation 1. This population produces thermodynamic trajectories $\omega$ of $K$ steps with the distribution $P(\eta)$ of efficiencies $\eta$ shown in \f{fig2}(a). Some of the networks in earlier generations do not appear in these distributions due to their undefined or very negative efficiencies. Even the best-performing members of this population produce efficiencies much lower than the Carnot efficiency, which is the most efficient trajectory possible given the set of allowed thermodynamic processes\c{Carnot}. We have $\eta_{\rm max} = 0.4$ with our choice of parameters.

We next perform the first step of evolutionary learning dynamics. We keep the 25 generation-1 networks whose trajectories have the largest $\eta$, and we discard the rest. We create 75 new networks by drawing uniformly from the set of 25, each time ``mutating'' all weights $w$ and biases $b$: for each weight or bias we draw a random number $\delta$ from a Gaussian distribution with zero mean and unit variance, and update the weight or bias as $w \to w + \epsilon \delta$ or $b \to b + \epsilon \delta$, where $\epsilon=0.05$ is an evolutionary learning rate. 

The new population of the 25 best generation-1 networks and their 75 mutant offspring constitute generation 2. We simulate those 100 networks for $K$ steps, producing the distribution of efficiencies shown in \f{fig2}(a). Continuing this alternation of evolutionary dynamics (retaining and mutating the best networks of the current generation) and physical dynamics (using the new generation of networks to generate a set of trajectories) gives rise to networks able to propagate increasingly efficient trajectories [\f{fig2}(a)]. After about 100 generations, we obtain networks whose efficiencies are equal to that of the Carnot cycle (to within four decimal places). Inspection of the trajectories corresponding to these values of $\eta$ show that they indeed form Carnot cycles; see \f{fig2}(b).

Several features of this learning process are notable. In learning to maximize the efficiency of a thermodynamic trajectory, networks have learned to propagate cycles, as opposed to non-closed trajectories in $P$-$V$ space, because cycles lead in general to larger efficiencies. As the thermal efficiency of the Carnot cycle, defined as
\begin{equation} \label{eq:carnot_eff}
\eta_{\rm max} = 1 - \frac{T_{\rm c}}{T_{\rm h}},
\end{equation}
is independent of volume, the Carnot cycle has no absolute scale associated with it, meaning the Carnot cycle on two heat engines with two distinct total volumes are equivalent. However, this is not true for the Stirling, Otto, or the discretized approximations of the Carnot cycles, whose thermal efficiencies can be simply derived using the \eqq{eq:eff_calc} with the processes in Tab. \ref{tab:action} \c{Callen_1985} and are defined as
\begin{equation}
\eta_{\rm S} = \frac{T_{\rm h} - T_{\rm c}}{T_{\rm h} + \frac{3\left(T_{\rm h} - T_{\rm c}\right)}{2\log\left(V_{r}\right)}},
\end{equation}
for the Stirling cycle, where $V_{r} = V_{\rm max}/V_{\rm min}$,
\begin{equation}
\eta_{\rm O} = 1 - V_{r}^{-\frac{2}{3}},    
\end{equation}
for the Otto cycle, and
\begin{equation}
\begin{aligned}
\eta_{\rm C}' &= \frac{T_{\rm h}\left(\log\left(\frac{V_{2}}{V_{\rm min}}\right) - \frac{3}{2}\left(\left(\frac{V_{2}}{V_{\rm max}}\right)^{\frac{2}{3}} - 1\right)\right)}{\frac{3}{2}\left(T_{\rm h} - T_{\rm c}\left(\frac{V_{1}}{V_{\rm min}}\right)^{\frac{2}{3}}\right) + T_{\rm h}\log\left(\frac{V_{2}}{V_{\rm min}}\right)} \\
&+ \frac{T_{\rm c}\left(\log\left(\frac{V_{1}}{V_{\rm max}}\right) - \frac{3}{2}\left(\left(\frac{V_{1}}{V_{\rm min}}\right)^{\frac{2}{3}} - 1\right)\right)}{\frac{3}{2}\left(T_{\rm h} - T_{\rm c}\left(\frac{V_{1}}{V_{\rm min}}\right)^{\frac{2}{3}}\right) + T_{\rm h}\log\left(\frac{V_{2}}{V_{\rm min}}\right)}
\end{aligned}
\end{equation}
for the discretized Carnot cycle, where $V_{1}$ is the volume when adiabatic compression begins, and $V_{2}$ is the volume when adiabatic expansion begins. If $V_{1} = (T_{\rm h} / T_{\rm c})^{3/2}V_{\rm min}$ and $V_{2} = (T_{\rm c} / T_{\rm h})^{3/2}V_{\rm max}$ then the volume terms cancel and we get $\eta_{\rm C}' = \eta_{\rm max}$. Otherwise, all of these equations contain volume terms and therefore the Stirling, Otto, and discretized Carnot cycles are volume dependent. 
Interestingly, $\eta_{\rm O}$ is independent of the thermal baths, however it can only be performed if $T_{\rm h} > T_{\rm c}V^{2/3}$. When this is the case we have
\begin{equation}
\begin{aligned}
\eta_{\rm max} &= 1 - \frac{T_{\rm c}}{T_{\rm h}} > 1 - \frac{T_{\rm c}}{T_{\rm c}V_{r}^{\frac{2}{3}}} = 1 - V_{r}^{-\frac{2}{3}} = \eta_{\rm O},
\end{aligned}
\end{equation}
however the Carnot cycle is not technically possible in this scenario as it is impossible to heat the system the required amount with adiabatic processes, so we only briefly consider it. For fixed $T_{\rm h}$ and $T_{\rm c}$, $\eta_{\rm S}$ and $\eta_{\rm O}$ are maximized by maximizing the volume ratio $V_{r}$, however $\eta_{\rm C}'$ is maximized by minimizing the difference between $V_{1}$ and $V_{2}$ with the volume values used in the true Carnot cycle. Given the fixed step sizes permitted for the processes in Tab. \ref{tab:action}, networks have learned to enact the size of a cycle that allows the best approximation of the Carnot cycle.

\begin{figure}
   \centering
   \includegraphics[width=\linewidth]{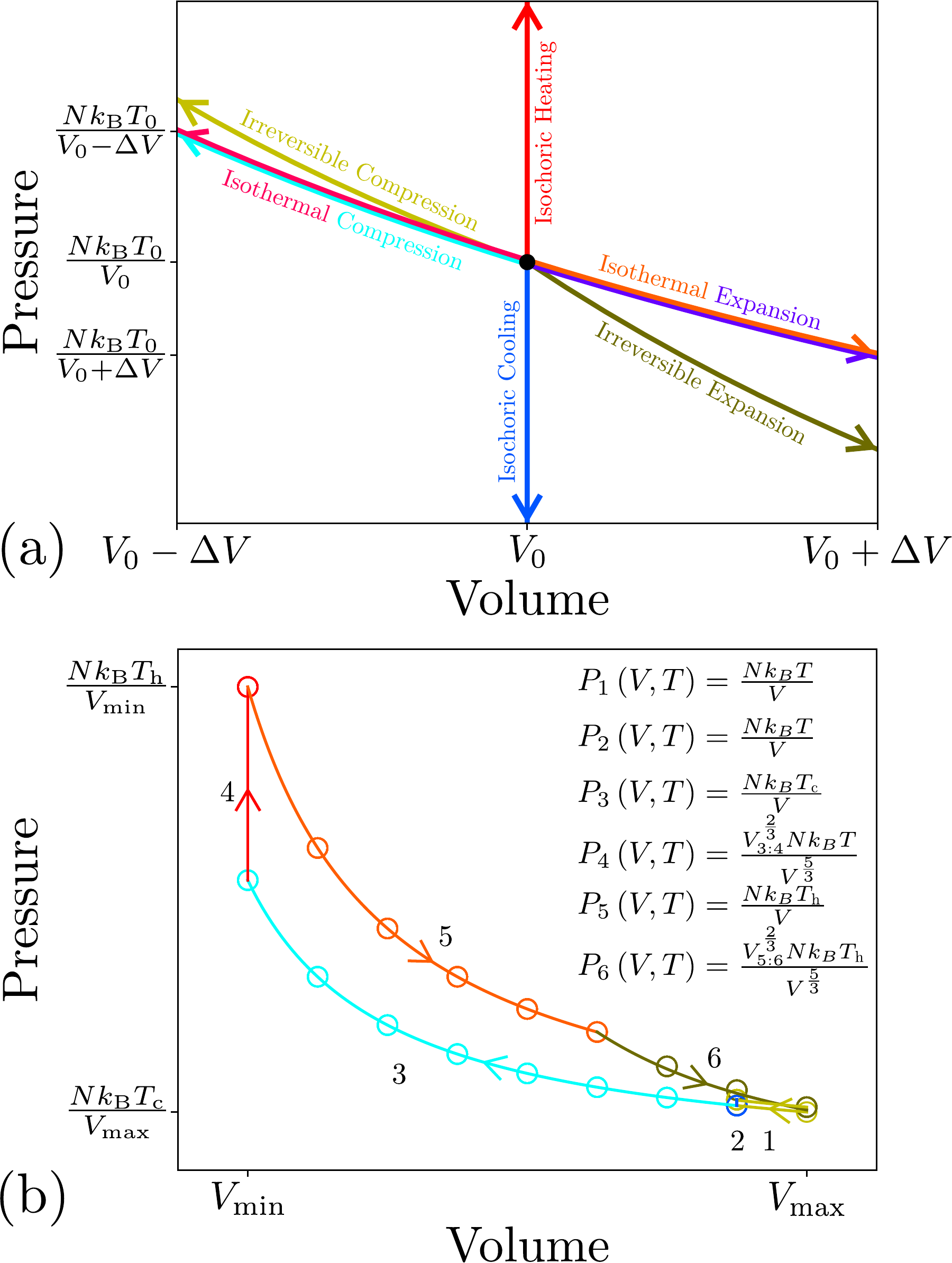} 
   \caption{We apply the evolutionary process described in \f{fig2} to a new setting in which the adiabatic processes of Tab. \ref{tab:action} are replaced with the irreversible process \eq{fict}; panel (a) summarizes the new set of accessible moves. (b) Highly-evolved networks learn to enact a hybrid of the Stirling and Carnot cycles, and the resulting equations of state can be identified by curve fitting.}
   \label{fig3}
\end{figure}

Given only a set of processes and a path-extensive measure of efficiency, our neural network-based evolutionary learning framework is able to maximize path efficiency and so deduce a classic result of physics. This learning framework is also successful if it is presented with a different set of processes. When denied the adiabatic processes of Tab. \ref{tab:action}, our gradient-free evolutionary algorithm learns the Stirling cycle\c{Stirling}, which is maximally-efficient in this context (with our systems parameters, $\eta_{\rm S} \approx 0.291$.); when denied the isothermal processes it similarly learns the maximally-efficient Otto cycle\c{Otto}. In fact, these are the only types of cycles possible in those scenarios. For the Otto cycle, $T_{\rm h}$, must be increased so we set it to 1000 K ($\eta_{\rm O} \approx 0.658$ and $\eta_{\rm max} = 0.8$).

Extensions to unknown thermodynamic processes are straightforward, and inspection of the resulting solutions provides physical insight in an unfamiliar setting. As an illustration, we replace the standard monatomic ideal gas adiabatic process, for which $TV^{2/3}$ is constant, with a fictitious irreversible process for which
\begin{equation}
\label{fict}
TV^{2/3} \propto (1-k)^{\Delta V/(V_0 - V_1)};
\end{equation}
here $k=2/5$ and $\Delta V$ are the fraction of thermal energy lost to the surrounding environment, and the change in volume upon making the move, respectively. We allow the network access to this process and the others of Tab. \ref{tab:action} (excluding adiabatic processes), summarized in \f{fig3}(a). In this setting, we do not know in advance the most efficient trajectory. In \f{fig3}(b) we show that the solution identified by our evolutionary learning scheme is a hybrid of the Stirling and Carnot cycles. Upon investigation of this solution, it is clear that the detriment to the thermal efficiency caused by the increased use of heat from the isochore (labelled process 4) is more beneficial than trying to approximate just the Carnot cycle given these irreversible processes. In this modified heat engine, we define an approximation of the Carnot cycle by replacing the adiabatic processes with the irreversible processes such that they heat and cool the system in a qualitatively similar manner. The irreversible compression process is thermally inefficient compared to adiabatic compression, therefore requiring more work to heat the system without using the thermal reservoirs. If $k$ is sufficiently low, the approximated Carnot cycle is most efficient and as $k$ approaches $1.0$, the hybrid cycle approaches the Stirling cycle. Note that for sufficiently high $k$ the irreversible process becomes physically unreasonable. By fitting equations to each branch of the cycle, we identify the equations of state that result from the irreversible process \eq{fict}. These results highlight the general applicability of the learning scheme and indicate the physical insight that can be obtained by interrogating solutions identified by machine learning.

\section{Gradient-based reinforcement learning}
As an alternative to the gradient-free evolutionary learning, we consider a method which includes gradient information. We note however, that while the observation $s_{t}=(V,T)$ contains sufficient information to correctly determine the required thermodynamic process at any step $t$, as shown above, it is not sufficient for determining thermal efficiency. For example, suppose the Carnot cycle requires $K$ steps with $s_{0}=(V_{\rm max},T_{\rm c})$. $K$ steps later we arrive back at the same observation, $s_{K}=s_{0}$, achieving a thermal efficiency of $\eta_{\rm max}$. Now if we perform the reverse Carnot cycle, we would begin and end at the same observations as before, however this time achieving a thermal efficiency of $-\eta_{\rm max}$, meaning that the relation of the current observation representation to thermal efficiency is one-to-many. Unlike gradient-free algorithms, traditional gradient-based reinforcement learning algorithms operate using the rewards associated with specific steps of a trajectory, instead of assigning a score to an entire trajectory. For this reason, these algorithms cannot be used on this specific simulated heat engine because it is not Markovian. A common example of a Markovian problem would be chess, where one needs to only know the current state of the board, not how the individual pieces were moved to their current positions in order to make an optimal move.

\subsection{Markovian model heat engine}
We now introduce a Markovian simulated heat engine designed for gradient-based reinforcement learning (we will also consider this Markovian system with our gradient-free approach). In this new problem, which will be referred to as the Markovian heat engine problem, we provide the agent with non-negative work and heat budgets, defined as $W^{*}_{K}= W_{0}^{*} + \sum_{t=0}^{K-1} \Delta W_{s_t s_{t+1}}$ and $Q^{*}_{K} = Q_{0}^{*} - \sum_{t=0}^{K-1} \Delta Q^{\rm in}_{s_{t} s_{t+1}}$ respectively, where $W_{0}^{*}$ and $Q_{0}^{*}$ are the initial budget values. For a thermodynamic process to be performed at time $t$, $W^{*}_{t-1} \geq -\Delta W_{s_{t-1} s_{t}}$ and $Q^{*}_{t-1} \geq Q^{\rm in}_{s_{t-1} s_{t}}$ are required to ensure $W^{*}_{t}$ and $Q^{*}_{t}$ are non-negative. These budgets represent the amount of work the agent can put into the system, where any work produced during a trajectory is added to the budget, and the amount of heat that can be transferred from the hot thermal reservoir to the heat engine. We define the observation for this new heat engine as $s_{t}=(V,T,W^{*}_{t},Q^{*}_{t})$,  however we still consider a cycle as a trajectory that returns the same values of $V$ and $T$. With this new representation, the observations at the beginning and end of a given singular cycle are now unique and Markovian. This means that a given observation contains all the required information about the state of the simulated heat engine and therefore the previous trajectory is not required. Full Markovity is not necessarily required to apply gradient-based reinforcement learning, however it does ensure the mapping from observation--action pairs to rewards can be learned by the agent. As $W_{0}^{*}$ and $Q_{0}^{*}$ are fixed, the relation of observation to thermal efficiency is now many-to-one, solving the issue of mapping an observation to a single thermal efficiency. 

As the observations for the start of each repeated cycle are unique, if a trajectory produces one cycle, it does not guarantee it will produce many cycles. This poses a greater challenge as each subsequent cycle must now be learned individually. This could be solved by re-initializing the system to the initial observation after a cycle is performed, however this would force a solution for single cycles where we would like to find a more general solution for the system. Thermal efficiency could be assigned to the final observation--action pair of a trajectory to update a given policy using gradients in order to produce the most thermally efficient cycle, similar to what was done in the previous case, however this does not encourage trajectories containing more than a single cycle as thermal efficiency does not change when a cycle is repeated exactly.

To avoid explicitly penalizing a trajectory for stopping after a single cycle, we instead use $\Delta W = W^{*}_{K} - W_{0}^{*}$ at the end of a trajectory. As $\Delta Q = Q_{0}^{*} - Q^{*}_{K}$ has a fixed finite maximum, using Eq. \ref{eq:eff_calc}, maximizing $\Delta W$ maximizes thermal efficiency while also encouraging trajectories with many cycles, satisfying our desired requirements. To help clarify this, consider the following example. Suppose some cycle on this heat engine produces 0.2 units of work and consumes 0.5 units of heat. This would give a thermal efficiency of 0.4. Now if we performed this cycle 10 times we would produce 2.0 units of work and consume 5.0 units of heat. However this would still give a thermal efficiency of 0.4. If we use thermal efficiency as the reward, these two cases are equivalent, however if we use $\Delta W$ as the reward, the second case is preferred.

\begin{figure*}
   \centering
   \includegraphics[width=\linewidth]{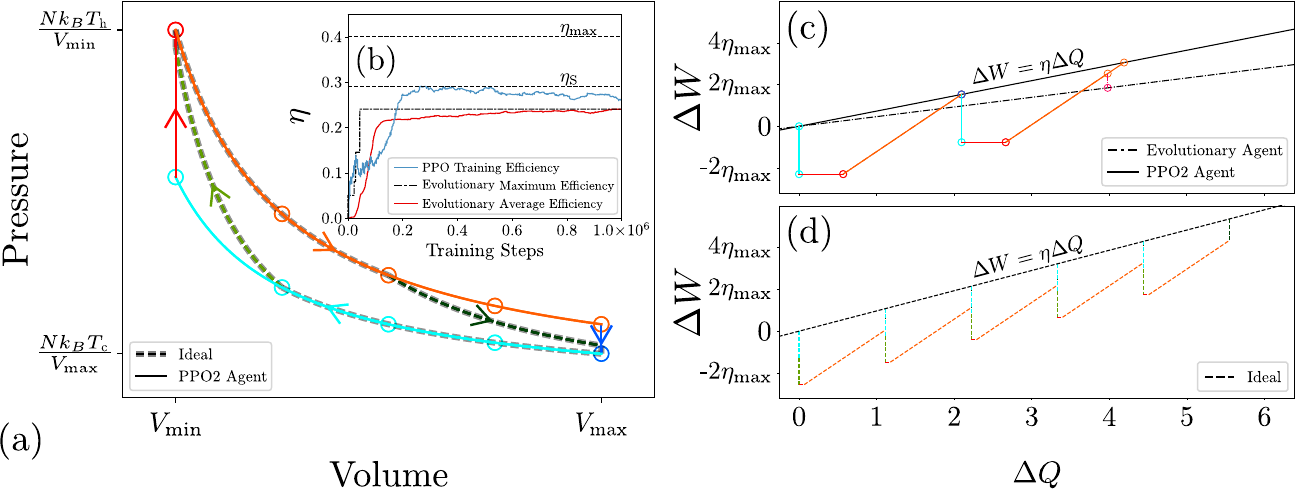} 
   \caption{We apply the PPO reinforcement learning process to our Markovian simulated heat engine; panel (a) shows the most thermally efficient trajectory produced in this process overlaid with the ideal solution, the discretized Carnot cycle. (b) The learning dynamics of thermal efficiency as a function of training steps for both the PPO and evolutionary  agents as they learn to maximize $\Delta W$. (c) $\Delta W$ as a function of $\Delta Q$ for the trajectory produced by the best performing PPO and evolutionary agents and (d) the ideal solution (discretized Carnot cycle), where $\Delta Q=1$ maps to the amount of energy used during a single analytic Carnot cycle.}
   \label{fig4}
\end{figure*}

\subsection{Proximal policy optimization}
With our newly defined engine and reward scheme, we turn to proximal policy optimization (PPO)\c{PPO} as the gradient-based reinforcement learning algorithm to find a policy, $\pi_{\theta}$, that produces the desired trajectories. PPO is a commonly used gradient-based reinforcement learning algorithm which has had good performance on many problems\c{bohn2019deep, melo2019learning, wei2019mixed, zhang2021image, de2021deep}. The true policy, $\pi_{\theta}^{\ast}$, requires certain features from the system in order to produce these desired trajectories. While the observation may not explicitly contain these features, we assume we can extract them using a neural network approximation of $\pi_{\theta}^{\ast}$. PPO aims to maximize the policy gradient objective function,
\begin{equation}
L^{\rm PG}_{t}\left(\theta\right) = \hat{\mathbb{E}}\left[\log \pi_{\theta}\left(a_{t} | s_{t}\right)\hat{A}_{t} \right],
\end{equation}
where the expectation is taken with respect to the actions $a_{t}$ and $\hat{A}_{t} = \hat{A}_{\theta}\left(s_{t} | a_{t}\right)$ is the estimate of the advantage function defined as the difference between the $\mathcal{Q}$-function $\mathcal{Q}_{\pi_{\theta}}(s_{t}, a_{t})$ and the value function $V_{\pi_{\theta}}(s_{t})$. The $\mathcal{Q}$-function is the expected discounted future reward if the process $a_{t}$ is performed starting at $s_{t}$ and $\pi_{\theta}$ is followed for the remainder of the trajectory starting at $s_{t+1}$, defined as
\begin{equation}
\mathcal{Q}_{\pi_{\theta}}(s_{t}, a_{t}) = r_{t} + \gamma \max_{a} \mathcal{Q}_{\pi_{\theta}}(s_{t+1}, a),
\end{equation}
where $0 \leq \gamma \leq 1$ is the discount factor. The value function is the expected discounted future reward if $\pi_{\theta}$ is followed for the remainder of the trajectory starting at $s_{t}$, defined as
\begin{equation}
\mathcal{V}_{\pi_{\theta}}(s_{t}) = \max_{a} \mathcal{Q}_{\pi_{\theta}}(s_{t}, a).
\end{equation}
Our estimate of future reward is rarely perfect, therefore it is discounted relative to the time scale of the reward. For example, when deciding on where to eat, we only care about the immediate reward we get, so we heavily discount our estimate of future reward in this case, however when investing in stocks, we care about the long-term reward, so our estimate of future reward should be minimally discounted. $\mathcal{Q}_{\pi_{\theta}}$ and $V_{\pi_{\theta}}$ are not known analytically but it is assumed they require the same features as $\pi_{\theta}$ and therefore we have our same neural network approximation of $\pi_{\theta}$ output the value at $s_{t}$, i.e. the output of the neural network is $(\pi_{\theta}(s_{t}), V_{\pi_{\theta}}(s_{t}))$. However, it is more efficient to instead maximize the trust region policy optimization (TRPO)\c{TRPO} objective function,
\begin{equation}
\begin{aligned}
L^{\rm CPI}_{t}\left(\theta\right) &= \hat{\mathbb{E}}\left[\frac{\pi_{\theta}\left(a_{t} | s_{t}\right)}{\pi_{\theta_{\rm old}}\left(a_{t} | s_{t}\right)}\hat{A}_{t}\right] \\
&= \hat{\mathbb{E}}\left[r_{\theta, \theta_{\rm old}}\left(a_{t} | s_{t}\right)\hat{A}_{t}\right],
\end{aligned}
\end{equation}
where $\pi_{\theta_{\rm old}}$ is the set of parameters of the previous policy. The trivial solution to this is to make extreme modifications to $\pi_{\theta}$, therefore PPO clips the probability ratio $r_{\theta, \theta_{\rm old}}$ and takes the minimum,
\begin{equation}
\begin{aligned}
L^{\rm CLIP}_{t}\left(\theta\right) &= \hat{\mathbb{E}}\left[\min\left(r_{\theta}\left(a_{t} | s_{t}\right)\hat{A}_{t}, r_{\theta}^{\rm CLIP}\left(a_{t} | s_{t}\right)\hat{A}_{t}\right)\right] \\
r_{\theta, \theta_{\rm old}}^{\rm CLIP}\left(a_{t} | s_{t}\right) &= {\rm clip}\left(r_{\theta, \theta_{\rm old}}\left(a_{t} | s_{t}\right), 1 - \epsilon, 1 + \epsilon\right),
\end{aligned}
\end{equation}
where $\epsilon$ is a hyperparameter and the clip function is defined as
\begin{equation}
{\rm clip}(x, a, b) = \min(\max(x, a), b) = \max(\min(x, b), a).
\end{equation}
While this objective function is designed to optimize the policy, it is not designed to optimize the value function, therefore we use a mean squared error cost function,
\begin{equation}
L^{\rm VF}_{t}\left(\theta\right) = \left(V_{\pi_{\theta}}\left(s_{t}\right) - V_{t}^{\rm targ}\right)^{2},
\end{equation}
where $V_{t}^{\rm targ}$ is the target value. Combing this cost function with our objective function and adding in a so called ``entropy'' term $S$ to encourage exploration, we have
\begin{equation}
L^{\rm PPO}_{t}\left(\theta\right) = \hat{\mathbb{E}}\left[L^{\rm CLIP}_{t}\left(\theta\right) - c_{1}L^{\rm VF}_{t}\left(\theta\right) + c_{2}S\left[\pi_{\theta}\right]\left(s_{t}\right)\right],
\end{equation}
for our overall PPO objective function, where $c_{1}$ and $c_{2}$ are hyper-parameters and $S\left[\pi_{\theta}\right]$ is defined by
\begin{equation}
S\left[\pi_{\theta}\right](s_{t}) = \sum_{a} \pi_{\theta}\left(a_{t} | s_{t}\right) \log\left(\pi_{\theta}\left(a_{t} | s_{t}\right)\right).
\end{equation}
Note that we have the negative of our cost term, so maximizing $L^{\rm PPO}_{t}$ simultaneously maximizes our objective function and minimizes our cost function. To update the parameters of our neural network, the gradient of $L^{\rm PPO}_{t}$ is estimated with respect to $\theta$ using backpropagation and we make changes to $\theta$ in the ascending direction of this gradient.

\subsection{Results and Discussion}
Using the identical network architecture as with the gradient-free method, we randomly initialize a policy and allow it to produce trajectories in the newly-defined thermodynamic system. As the policy interacts with this system, we collect tuples of the form $(s_{t}, a_{t}, r_{t}, s_{t+1})$ which are the experiences used to update the policy with PPO.  We require all four of these because the reward $r_{t}$ specifically arises from how the action $a_{t}$ maps the observation $s_{t}$ to the next observation $s_{t+1}$. $r_{t}$ and $s_{t+1}$ are also required to update our approximation of the value function and calculate $\hat{A}_{t}$. To perform these updates, we used the following hyper-parameters: discount factor of 0.99, $\epsilon=0.2$, $c_{1}=0.5$, $c_{2}=0.01$, a batch size of 128, a total number of training steps of $2 \times 10^{6}$, and a learning rate of $2.5 \times 10^{-4}$ which determines the step size when updating the parameters $\theta$.

Unlike the gradient-free approach previously discussed, PPO is unable to achieve an optimal result. However the agent was able to produce a trajectory that not only follows the Stirling cycle as shown in \f{fig4}(a), and does so more than once. When looking at the learning dynamics shown in \f{fig4}(b), we can see that the policy that produces the most thermally efficient trajectory occurs at $\sim 2.5 \times 10^{5}$ training steps, and then thermal efficiency steadily decreases in all subsequent policies. This is likely due to the Stirling cycle being a local maximum (in terms of $\Delta W$). When a policy is at a local maximum for long enough, the value function is optimized on the trajectories found at this local maximum to the point that it becomes useless when evaluated on any trajectories found outside of the local maximum. Due to the deterministic nature of these problems, only a single trajectory is seen at this local maximum. Eventually, the entropy term used for stochastic exploration forces the policy to stray from this locally optimal trajectory, causing the policy to produce a trajectory unseen by the agent recently. As the value function is no longer accurate at this observation, the agent is unable to bring the policy back to this local maximum causing performance to diminish. The Stirling cycle is a local maximum because it produces more work than the Carnot cycle, i.e. achieves a higher reward per cycle, however as seen when comparing \ff{fig4}(c) and (d), the Stirling cycle consumes a greater amount of heat than the Carnot cycle. This makes it more viable in the short term, however, due to the upper bound constraint placed on $\Delta Q$, the Carnot cycle is still the ideal solution when tasked with maximizing $\Delta W$. 

With the maximum amount of heat allowed, the only way to increase $\Delta W$ beyond what is produced by two Stirling cycles, a trajectory would need to instead produce either three Carnot cycles and one Stirling cycle or five Carnot cycles. The differences between these trajectories and the one the agent produces are not trivial. Making any slight modifications to the found trajectory decreases $\Delta W$. It is only when an entire Stirling cycle is replaced with multiple Carnot cycles in a trajectory that there would be any improvement. Additionally, the reward signal used to update a policy comes from the end of a trajectory, whereas due to the repetitive nature of the non-Markovian problem, we were able to select the maximal score along the entire trajectory. Now the policy is required to produce trajectories that not only follow optimal paths but also end in optimal positions, increasing the difficulty of the Markovian problem even further.

\begin{figure*}
   \centering
   \includegraphics[width=\linewidth]{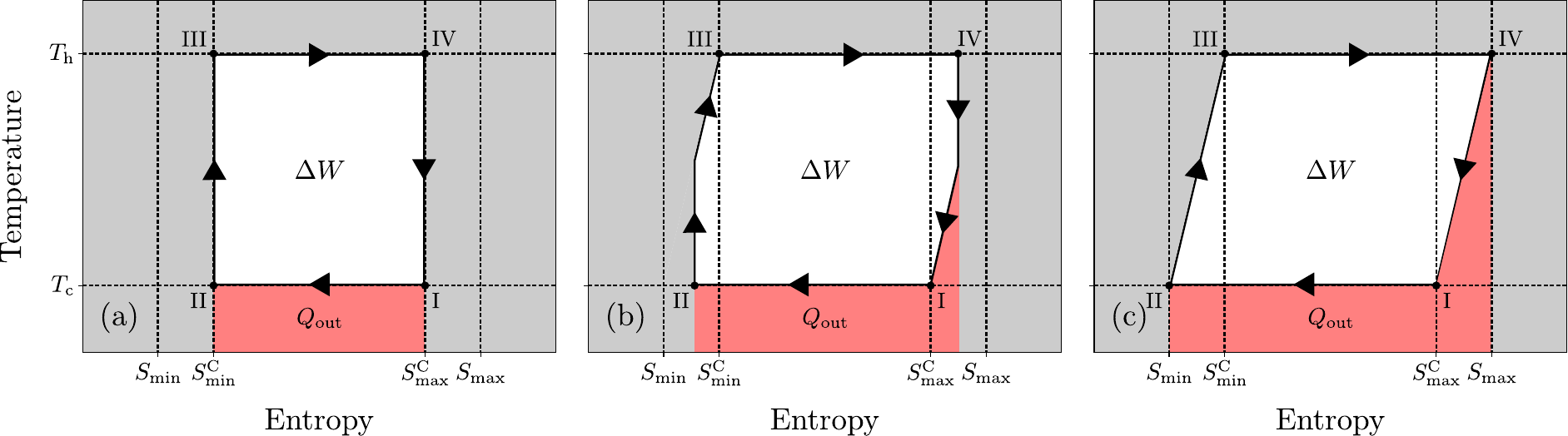}
   \caption{$T$-$S$ diagram for the (a) Carnot, (b) discretized Carnot, and (c) Stirling cycles. The red (dark gray) area (labelled $Q_{\rm out}$) represents the heat removed from the system, the white area (labelled $\Delta W$) represents the work produced by the system, and the red (dark gray) and white areas together represent the heat added to the system.}
   \label{fig5}
\end{figure*}

\section{Evolutionary learning on Markovian heat engine}
If $\Delta W$ is used as the scoring metric, the evolutionary learning method from before can also be applied to this variant of the simulated heat engine. Doing so yields trajectories with {\em lower} thermal efficiencies and less work produced compared with the gradient-based method as shown in \ff{fig4}(b) and (c). Similar to the gradient-based method, the gradient-free method starts it's trajectory by producing a Stirling cycle, however it is unable to complete the second cycle. While the desired trajectory is identical to that of the non-Markovian problem, this task is a more challenging one because the cycle must be learned over and over again as the heat budget is consumed. This shows that the evolutionary learning method is not necessarily superior to the gradient-based method in all ways, but rather each approach is suited for different tasks. The gradient-based method was able to outperform the evolutionary learning method on the Markovian problem, however, it was the evolutionary learning method that was able to find the maximally efficient trajectory when applied to the non-Markovian problem. The evolutionary learning method was able to be applied to both tasks, however, because of the nature of gradient-based reinforcement learning, we could only apply it to the Markovian problem. The fact the gradient-based method finds this local maximum while the gradient-free does not supports the idea that gradient-based reinforcement learning can propagate the reward assigned to specific observation--action pairs in order to produce at least a partially optimal solution. Evolutionary learning methods cannot straightforwardly make use of these specific assignments and this is likely the reason why it is outperformed by the gradient-based learning method on the Markovian problem. Although the evolutionary learning method seems to be the go-to method, being able to define a problem the way we did for the non-Markovian one is not always possible. It would be possible to make the Markovian problem easier for the gradient-based learning method, but not without providing such a high level of feedback that the problem becomes pointless to solve. For example, one could reward the gradient-based learning agent as it performs each of the correct steps in the most thermally efficient trajectory, however, this approach would only work in the case an optimal solution was known beforehand. Novelty search techniques might allow this gradient-based learning method to find the optimal trajectory, however, this would increase the computational cost of this method even further. Given that gradient-based methods are already more computationally expensive than gradient-free methods, and that the gradient-free method has already achieved the optimal result, we have chosen not to explore these options.

\section{Conclusions}
Motivated by the correspondence between games and physical trajectories, we have shown that neural network-based evolutionary learning can optimize the efficiency of trajectories of classical thermodynamics. Given a set of physical processes and a path-extensive measure of efficiency, networks evolve to learn the maximally-efficient Carnot, Stirling, or Otto cycles, reproducing classic results of physics that were originally derived by application of physical insight. Given new processes, the evolutionary framework identifies solutions that when interrogated provide physical insight into the problem at hand. We have also shown that with slight modifications, gradient-based reinforcement learning can optimize the efficiency of trajectories of classical thermodynamics, although not quite as effectively.

\begin{acknowledgments}
CB, UY, RC, KM, and IT performed work at the National Research Council of Canada under the auspices of the AI4D Program. IT acknowledges NSERC. SW performed work at the Molecular Foundry, Lawrence Berkeley National Laboratory, supported by the Office of Science, Office of Basic Energy Sciences, of the U.S. Department of Energy under Contract No. DE-AC02--05CH11231.
\end{acknowledgments}

The simulated heat engines (or environments\c{OpenAI} as commonly referred to in the reinforcement learning community) used in this study can be found at \url{https://clean.energyscience.ca/gyms}.

\appendix
\section{Proof of Carnot's theorem on infinite heat bathes}
\label{sec:proof}
Using $T$-$S$ diagrams, shown in \f{fig5}, the thermal efficiency of a cycle is determined by $\eta = \Delta W / (\Delta W + Q_{\rm out})$ where $\Delta W$ is the work produced by the system and $Q_{\rm out}$ is the heat removed from the system. Using \f{fig5}(a), the thermal efficiency of the Carnot cycle is given by \eqq{eq:carnot_eff}, which is independent of the minimum and maximum entropies reach by the Carnot cycle, denoted by $S_{\rm min}^{\rm C}$ and $S_{\rm max}^{\rm C}$ respectively. For cycles that operate between $S_{\rm min}^{\rm C}$ and $S_{\rm max}^{\rm C}$, since $T_{\rm c}$ is the minimum temperature the system can reach, $\eta$ is maximized by maximizing $\Delta W$ and minimizing $Q_{\rm out}$, which results in the Carnot cycle. Now consider cycles that operate outside of the entropy range $S_{\rm min}^{\rm C}$ to $S_{\rm max}^{\rm C}$, such as the ones shown in \ff{fig5}(b) and (c). It is impossible for the system to simultaneously be at $T_{\rm h}$ and an entropy less than $S_{\rm min}^{\rm C}$, therefore the ratio of work produced and heat removed by the system between points II and III will be less than that of the Carnot cycle. Similarly, it is impossible for the system to simultaneously be at $T_{\rm c}$ and and entropy greater than $S_{\rm max}^{\rm C}$, therefore the ratio of work produced and heat removed by the system between points IV and I will be less than the Carnot cycle. Putting all these together, any cycle operating within $S_{\rm min}^{\rm C}$ and $S_{\rm max}^{\rm C}$ is at most efficient as the Carnot cycle, and any cycle operating outside of $S_{\rm min}^{\rm C}$ and $S_{\rm max}^{\rm C}$ is less efficient, therefore Carnot's theorem holds for infinitely many heat bathes.

\bibliography{main.bib}

\begin{thebibliography}{52}
\expandafter\ifx\csname natexlab\endcsname\relax\def\natexlab#1{#1}\fi
\expandafter\ifx\csname bibnamefont\endcsname\relax
  \def\bibnamefont#1{#1}\fi
\expandafter\ifx\csname bibfnamefont\endcsname\relax
  \def\bibfnamefont#1{#1}\fi
\expandafter\ifx\csname citenamefont\endcsname\relax
  \def\citenamefont#1{#1}\fi
\expandafter\ifx\csname url\endcsname\relax
  \def\url#1{\texttt{#1}}\fi
\expandafter\ifx\csname urlprefix\endcsname\relax\def\urlprefix{URL }\fi
\providecommand{\bibinfo}[2]{#2}
\providecommand{\eprint}[2][]{\url{#2}}

\bibitem[{\citenamefont{Roberts et~al.}(1959)\citenamefont{Roberts, Arth, and
  Bush}}]{roberts1959games}
\bibinfo{author}{\bibfnamefont{J.~M.} \bibnamefont{Roberts}},
  \bibinfo{author}{\bibfnamefont{M.~J.} \bibnamefont{Arth}}, \bibnamefont{and}
  \bibinfo{author}{\bibfnamefont{R.~R.} \bibnamefont{Bush}},
  \bibinfo{journal}{American anthropologist} \textbf{\bibinfo{volume}{61}},
  \bibinfo{pages}{597} (\bibinfo{year}{1959}).

\bibitem[{\citenamefont{Watkins and Dayan}(1992)}]{QL}
\bibinfo{author}{\bibfnamefont{C.~J.} \bibnamefont{Watkins}} \bibnamefont{and}
  \bibinfo{author}{\bibfnamefont{P.}~\bibnamefont{Dayan}},
  \bibinfo{journal}{Machine learning} \textbf{\bibinfo{volume}{8}},
  \bibinfo{pages}{279} (\bibinfo{year}{1992}).

\bibitem[{\citenamefont{Mnih et~al.}(2013)\citenamefont{Mnih, Kavukcuoglu,
  Silver, Graves, Antonoglou, Wierstra, and Riedmiller}}]{DQN}
\bibinfo{author}{\bibfnamefont{V.}~\bibnamefont{Mnih}},
  \bibinfo{author}{\bibfnamefont{K.}~\bibnamefont{Kavukcuoglu}},
  \bibinfo{author}{\bibfnamefont{D.}~\bibnamefont{Silver}},
  \bibinfo{author}{\bibfnamefont{A.}~\bibnamefont{Graves}},
  \bibinfo{author}{\bibfnamefont{I.}~\bibnamefont{Antonoglou}},
  \bibinfo{author}{\bibfnamefont{D.}~\bibnamefont{Wierstra}}, \bibnamefont{and}
  \bibinfo{author}{\bibfnamefont{M.}~\bibnamefont{Riedmiller}},
  \bibinfo{journal}{arXiv preprint arXiv:1312.5602}  (\bibinfo{year}{2013}).

\bibitem[{\citenamefont{Mnih et~al.}(2015)\citenamefont{Mnih, Kavukcuoglu,
  Silver, Rusu, Veness, Bellemare, Graves, Riedmiller, Fidjeland, Ostrovski
  et~al.}}]{Atari}
\bibinfo{author}{\bibfnamefont{V.}~\bibnamefont{Mnih}},
  \bibinfo{author}{\bibfnamefont{K.}~\bibnamefont{Kavukcuoglu}},
  \bibinfo{author}{\bibfnamefont{D.}~\bibnamefont{Silver}},
  \bibinfo{author}{\bibfnamefont{A.~A.} \bibnamefont{Rusu}},
  \bibinfo{author}{\bibfnamefont{J.}~\bibnamefont{Veness}},
  \bibinfo{author}{\bibfnamefont{M.~G.} \bibnamefont{Bellemare}},
  \bibinfo{author}{\bibfnamefont{A.}~\bibnamefont{Graves}},
  \bibinfo{author}{\bibfnamefont{M.}~\bibnamefont{Riedmiller}},
  \bibinfo{author}{\bibfnamefont{A.~K.} \bibnamefont{Fidjeland}},
  \bibinfo{author}{\bibfnamefont{G.}~\bibnamefont{Ostrovski}},
  \bibnamefont{et~al.}, \bibinfo{journal}{Nature}
  \textbf{\bibinfo{volume}{518}}, \bibinfo{pages}{529} (\bibinfo{year}{2015}).

\bibitem[{\citenamefont{Bellemare et~al.}(2013)\citenamefont{Bellemare, Naddaf,
  Veness, and Bowling}}]{Atari2600}
\bibinfo{author}{\bibfnamefont{M.~G.} \bibnamefont{Bellemare}},
  \bibinfo{author}{\bibfnamefont{Y.}~\bibnamefont{Naddaf}},
  \bibinfo{author}{\bibfnamefont{J.}~\bibnamefont{Veness}}, \bibnamefont{and}
  \bibinfo{author}{\bibfnamefont{M.}~\bibnamefont{Bowling}},
  \bibinfo{journal}{Journal of Artificial Intelligence Research}
  \textbf{\bibinfo{volume}{47}}, \bibinfo{pages}{253} (\bibinfo{year}{2013}).

\bibitem[{\citenamefont{Mnih et~al.}(2016)\citenamefont{Mnih, Badia, Mirza,
  Graves, Lillicrap, Harley, Silver, and Kavukcuoglu}}]{Actor}
\bibinfo{author}{\bibfnamefont{V.}~\bibnamefont{Mnih}},
  \bibinfo{author}{\bibfnamefont{A.~P.} \bibnamefont{Badia}},
  \bibinfo{author}{\bibfnamefont{M.}~\bibnamefont{Mirza}},
  \bibinfo{author}{\bibfnamefont{A.}~\bibnamefont{Graves}},
  \bibinfo{author}{\bibfnamefont{T.}~\bibnamefont{Lillicrap}},
  \bibinfo{author}{\bibfnamefont{T.}~\bibnamefont{Harley}},
  \bibinfo{author}{\bibfnamefont{D.}~\bibnamefont{Silver}}, \bibnamefont{and}
  \bibinfo{author}{\bibfnamefont{K.}~\bibnamefont{Kavukcuoglu}}, in
  \emph{\bibinfo{booktitle}{International conference on machine learning}}
  (\bibinfo{year}{2016}), pp. \bibinfo{pages}{1928--1937}.

\bibitem[{\citenamefont{Tassa et~al.}(2018)\citenamefont{Tassa, Doron, Muldal,
  Erez, Li, Casas, Budden, Abdolmaleki, Merel, Lefrancq et~al.}}]{DeepMind}
\bibinfo{author}{\bibfnamefont{Y.}~\bibnamefont{Tassa}},
  \bibinfo{author}{\bibfnamefont{Y.}~\bibnamefont{Doron}},
  \bibinfo{author}{\bibfnamefont{A.}~\bibnamefont{Muldal}},
  \bibinfo{author}{\bibfnamefont{T.}~\bibnamefont{Erez}},
  \bibinfo{author}{\bibfnamefont{Y.}~\bibnamefont{Li}},
  \bibinfo{author}{\bibfnamefont{D.~d.~L.} \bibnamefont{Casas}},
  \bibinfo{author}{\bibfnamefont{D.}~\bibnamefont{Budden}},
  \bibinfo{author}{\bibfnamefont{A.}~\bibnamefont{Abdolmaleki}},
  \bibinfo{author}{\bibfnamefont{J.}~\bibnamefont{Merel}},
  \bibinfo{author}{\bibfnamefont{A.}~\bibnamefont{Lefrancq}},
  \bibnamefont{et~al.}, \bibinfo{journal}{arXiv preprint arXiv:1801.00690}
  (\bibinfo{year}{2018}).

\bibitem[{\citenamefont{Todorov et~al.}(2012)\citenamefont{Todorov, Erez, and
  Tassa}}]{MuJoCo}
\bibinfo{author}{\bibfnamefont{E.}~\bibnamefont{Todorov}},
  \bibinfo{author}{\bibfnamefont{T.}~\bibnamefont{Erez}}, \bibnamefont{and}
  \bibinfo{author}{\bibfnamefont{Y.}~\bibnamefont{Tassa}}, in
  \emph{\bibinfo{booktitle}{Intelligent Robots and Systems (IROS), 2012
  IEEE/RSJ International Conference on}} (\bibinfo{organization}{IEEE},
  \bibinfo{year}{2012}), pp. \bibinfo{pages}{5026--5033}.

\bibitem[{\citenamefont{Puterman}(2014)}]{MDP}
\bibinfo{author}{\bibfnamefont{M.~L.} \bibnamefont{Puterman}},
  \emph{\bibinfo{title}{Markov decision processes: discrete stochastic dynamic
  programming}} (\bibinfo{publisher}{John Wiley \& Sons},
  \bibinfo{year}{2014}).

\bibitem[{\citenamefont{Asperti et~al.}(2019)\citenamefont{Asperti, Cortesi,
  and Sovrano}}]{Rogue}
\bibinfo{author}{\bibfnamefont{A.}~\bibnamefont{Asperti}},
  \bibinfo{author}{\bibfnamefont{D.}~\bibnamefont{Cortesi}}, \bibnamefont{and}
  \bibinfo{author}{\bibfnamefont{F.}~\bibnamefont{Sovrano}}, in
  \emph{\bibinfo{booktitle}{Machine Learning, Optimization, and Data Science}},
  edited by \bibinfo{editor}{\bibfnamefont{G.}~\bibnamefont{Nicosia}},
  \bibinfo{editor}{\bibfnamefont{P.}~\bibnamefont{Pardalos}},
  \bibinfo{editor}{\bibfnamefont{G.}~\bibnamefont{Giuffrida}},
  \bibinfo{editor}{\bibfnamefont{R.}~\bibnamefont{Umeton}}, \bibnamefont{and}
  \bibinfo{editor}{\bibfnamefont{V.}~\bibnamefont{Sciacca}}
  (\bibinfo{publisher}{Springer International Publishing},
  \bibinfo{address}{Cham}, \bibinfo{year}{2019}), pp.
  \bibinfo{pages}{264--275}.

\bibitem[{\citenamefont{Riedmiller}(2005)}]{NFQ}
\bibinfo{author}{\bibfnamefont{M.}~\bibnamefont{Riedmiller}}, in
  \emph{\bibinfo{booktitle}{European Conference on Machine Learning}}
  (\bibinfo{organization}{Springer}, \bibinfo{year}{2005}), pp.
  \bibinfo{pages}{317--328}.

\bibitem[{\citenamefont{Riedmiller et~al.}(2009)\citenamefont{Riedmiller,
  Gabel, Hafner, and Lange}}]{Soccer}
\bibinfo{author}{\bibfnamefont{M.}~\bibnamefont{Riedmiller}},
  \bibinfo{author}{\bibfnamefont{T.}~\bibnamefont{Gabel}},
  \bibinfo{author}{\bibfnamefont{R.}~\bibnamefont{Hafner}}, \bibnamefont{and}
  \bibinfo{author}{\bibfnamefont{S.}~\bibnamefont{Lange}},
  \bibinfo{journal}{Autonomous Robots} \textbf{\bibinfo{volume}{27}},
  \bibinfo{pages}{55} (\bibinfo{year}{2009}).

\bibitem[{\citenamefont{Schulman et~al.}(2017)\citenamefont{Schulman, Wolski,
  Dhariwal, Radford, and Klimov}}]{PPO}
\bibinfo{author}{\bibfnamefont{J.}~\bibnamefont{Schulman}},
  \bibinfo{author}{\bibfnamefont{F.}~\bibnamefont{Wolski}},
  \bibinfo{author}{\bibfnamefont{P.}~\bibnamefont{Dhariwal}},
  \bibinfo{author}{\bibfnamefont{A.}~\bibnamefont{Radford}}, \bibnamefont{and}
  \bibinfo{author}{\bibfnamefont{O.}~\bibnamefont{Klimov}},
  \bibinfo{journal}{arXiv preprint arXiv:1707.06347}  (\bibinfo{year}{2017}).

\bibitem[{\citenamefont{Such et~al.}(2017)\citenamefont{Such, Madhavan, Conti,
  Lehman, Stanley, and Clune}}]{Guber}
\bibinfo{author}{\bibfnamefont{F.~P.} \bibnamefont{Such}},
  \bibinfo{author}{\bibfnamefont{V.}~\bibnamefont{Madhavan}},
  \bibinfo{author}{\bibfnamefont{E.}~\bibnamefont{Conti}},
  \bibinfo{author}{\bibfnamefont{J.}~\bibnamefont{Lehman}},
  \bibinfo{author}{\bibfnamefont{K.~O.} \bibnamefont{Stanley}},
  \bibnamefont{and} \bibinfo{author}{\bibfnamefont{J.}~\bibnamefont{Clune}},
  \bibinfo{journal}{arXiv preprint arXiv:1712.06567}  (\bibinfo{year}{2017}).

\bibitem[{\citenamefont{Brockman et~al.}(2016)\citenamefont{Brockman, Cheung,
  Pettersson, Schneider, Schulman, Tang, and Zaremba}}]{OpenAI}
\bibinfo{author}{\bibfnamefont{G.}~\bibnamefont{Brockman}},
  \bibinfo{author}{\bibfnamefont{V.}~\bibnamefont{Cheung}},
  \bibinfo{author}{\bibfnamefont{L.}~\bibnamefont{Pettersson}},
  \bibinfo{author}{\bibfnamefont{J.}~\bibnamefont{Schneider}},
  \bibinfo{author}{\bibfnamefont{J.}~\bibnamefont{Schulman}},
  \bibinfo{author}{\bibfnamefont{J.}~\bibnamefont{Tang}}, \bibnamefont{and}
  \bibinfo{author}{\bibfnamefont{W.}~\bibnamefont{Zaremba}},
  \bibinfo{journal}{arXiv preprint arXiv:1606.01540}  (\bibinfo{year}{2016}).

\bibitem[{\citenamefont{Kempka et~al.}(2016)\citenamefont{Kempka, Wydmuch,
  Runc, Toczek, and Ja{\'s}kowski}}]{VizDoom}
\bibinfo{author}{\bibfnamefont{M.}~\bibnamefont{Kempka}},
  \bibinfo{author}{\bibfnamefont{M.}~\bibnamefont{Wydmuch}},
  \bibinfo{author}{\bibfnamefont{G.}~\bibnamefont{Runc}},
  \bibinfo{author}{\bibfnamefont{J.}~\bibnamefont{Toczek}}, \bibnamefont{and}
  \bibinfo{author}{\bibfnamefont{W.}~\bibnamefont{Ja{\'s}kowski}}, in
  \emph{\bibinfo{booktitle}{Computational Intelligence and Games (CIG), 2016
  IEEE Conference on}} (\bibinfo{organization}{IEEE}, \bibinfo{year}{2016}),
  pp. \bibinfo{pages}{1--8}.

\bibitem[{\citenamefont{{Wydmuch} et~al.}(2019)\citenamefont{{Wydmuch},
  {Kempka}, and {Jaśkowski}}}]{VizDoom2}
\bibinfo{author}{\bibfnamefont{M.}~\bibnamefont{{Wydmuch}}},
  \bibinfo{author}{\bibfnamefont{M.}~\bibnamefont{{Kempka}}}, \bibnamefont{and}
  \bibinfo{author}{\bibfnamefont{W.}~\bibnamefont{{Jaśkowski}}},
  \bibinfo{journal}{IEEE Transactions on Games} \textbf{\bibinfo{volume}{11}},
  \bibinfo{pages}{248} (\bibinfo{year}{2019}).

\bibitem[{\citenamefont{Silver et~al.}(2016)\citenamefont{Silver, Huang,
  Maddison, Guez, Sifre, Van Den~Driessche, Schrittwieser, Antonoglou,
  Panneershelvam, Lanctot et~al.}}]{Go}
\bibinfo{author}{\bibfnamefont{D.}~\bibnamefont{Silver}},
  \bibinfo{author}{\bibfnamefont{A.}~\bibnamefont{Huang}},
  \bibinfo{author}{\bibfnamefont{C.~J.} \bibnamefont{Maddison}},
  \bibinfo{author}{\bibfnamefont{A.}~\bibnamefont{Guez}},
  \bibinfo{author}{\bibfnamefont{L.}~\bibnamefont{Sifre}},
  \bibinfo{author}{\bibfnamefont{G.}~\bibnamefont{Van Den~Driessche}},
  \bibinfo{author}{\bibfnamefont{J.}~\bibnamefont{Schrittwieser}},
  \bibinfo{author}{\bibfnamefont{I.}~\bibnamefont{Antonoglou}},
  \bibinfo{author}{\bibfnamefont{V.}~\bibnamefont{Panneershelvam}},
  \bibinfo{author}{\bibfnamefont{M.}~\bibnamefont{Lanctot}},
  \bibnamefont{et~al.}, \bibinfo{journal}{nature}
  \textbf{\bibinfo{volume}{529}}, \bibinfo{pages}{484} (\bibinfo{year}{2016}).

\bibitem[{\citenamefont{Silver et~al.}(2017)\citenamefont{Silver,
  Schrittwieser, Simonyan, Antonoglou, Huang, Guez, Hubert, Baker, Lai, Bolton
  et~al.}}]{Go2}
\bibinfo{author}{\bibfnamefont{D.}~\bibnamefont{Silver}},
  \bibinfo{author}{\bibfnamefont{J.}~\bibnamefont{Schrittwieser}},
  \bibinfo{author}{\bibfnamefont{K.}~\bibnamefont{Simonyan}},
  \bibinfo{author}{\bibfnamefont{I.}~\bibnamefont{Antonoglou}},
  \bibinfo{author}{\bibfnamefont{A.}~\bibnamefont{Huang}},
  \bibinfo{author}{\bibfnamefont{A.}~\bibnamefont{Guez}},
  \bibinfo{author}{\bibfnamefont{T.}~\bibnamefont{Hubert}},
  \bibinfo{author}{\bibfnamefont{L.}~\bibnamefont{Baker}},
  \bibinfo{author}{\bibfnamefont{M.}~\bibnamefont{Lai}},
  \bibinfo{author}{\bibfnamefont{A.}~\bibnamefont{Bolton}},
  \bibnamefont{et~al.}, \bibinfo{journal}{Nature}
  \textbf{\bibinfo{volume}{550}}, \bibinfo{pages}{354} (\bibinfo{year}{2017}).

\bibitem[{\citenamefont{Mills et~al.}(2020)\citenamefont{Mills, Ronagh, and
  Tamblyn}}]{mills2020finding}
\bibinfo{author}{\bibfnamefont{K.}~\bibnamefont{Mills}},
  \bibinfo{author}{\bibfnamefont{P.}~\bibnamefont{Ronagh}}, \bibnamefont{and}
  \bibinfo{author}{\bibfnamefont{I.}~\bibnamefont{Tamblyn}},
  \bibinfo{journal}{Nature Machine Intelligence} \textbf{\bibinfo{volume}{2}},
  \bibinfo{pages}{509} (\bibinfo{year}{2020}).

\bibitem[{\citenamefont{Andreasson et~al.}(2019)\citenamefont{Andreasson,
  Johansson, Liljestrand, and Granath}}]{andreasson2019quantum}
\bibinfo{author}{\bibfnamefont{P.}~\bibnamefont{Andreasson}},
  \bibinfo{author}{\bibfnamefont{J.}~\bibnamefont{Johansson}},
  \bibinfo{author}{\bibfnamefont{S.}~\bibnamefont{Liljestrand}},
  \bibnamefont{and} \bibinfo{author}{\bibfnamefont{M.}~\bibnamefont{Granath}},
  \bibinfo{journal}{Quantum} \textbf{\bibinfo{volume}{3}}, \bibinfo{pages}{183}
  (\bibinfo{year}{2019}).

\bibitem[{\citenamefont{Zhou et~al.}(2019)\citenamefont{Zhou, Kearnes, Li,
  Zare, and Riley}}]{zhou2019optimization}
\bibinfo{author}{\bibfnamefont{Z.}~\bibnamefont{Zhou}},
  \bibinfo{author}{\bibfnamefont{S.}~\bibnamefont{Kearnes}},
  \bibinfo{author}{\bibfnamefont{L.}~\bibnamefont{Li}},
  \bibinfo{author}{\bibfnamefont{R.~N.} \bibnamefont{Zare}}, \bibnamefont{and}
  \bibinfo{author}{\bibfnamefont{P.}~\bibnamefont{Riley}},
  \bibinfo{journal}{Scientific reports} \textbf{\bibinfo{volume}{9}},
  \bibinfo{pages}{1} (\bibinfo{year}{2019}).

\bibitem[{\citenamefont{Radaideh et~al.}(2020)\citenamefont{Radaideh,
  Wolverton, Joseph, Tusar, Otgonbaatar, Roy, Forget, and
  Shirvan}}]{radaideh2020physics}
\bibinfo{author}{\bibfnamefont{M.~I.} \bibnamefont{Radaideh}},
  \bibinfo{author}{\bibfnamefont{I.}~\bibnamefont{Wolverton}},
  \bibinfo{author}{\bibfnamefont{J.}~\bibnamefont{Joseph}},
  \bibinfo{author}{\bibfnamefont{J.~J.} \bibnamefont{Tusar}},
  \bibinfo{author}{\bibfnamefont{U.}~\bibnamefont{Otgonbaatar}},
  \bibinfo{author}{\bibfnamefont{N.}~\bibnamefont{Roy}},
  \bibinfo{author}{\bibfnamefont{B.}~\bibnamefont{Forget}}, \bibnamefont{and}
  \bibinfo{author}{\bibfnamefont{K.}~\bibnamefont{Shirvan}},
  \bibinfo{journal}{Nuclear Engineering and Design} p. \bibinfo{pages}{110966}
  (\bibinfo{year}{2020}).

\bibitem[{\citenamefont{Badloe et~al.}(2020)\citenamefont{Badloe, Kim, and
  Rho}}]{badloe2020biomimetic}
\bibinfo{author}{\bibfnamefont{T.}~\bibnamefont{Badloe}},
  \bibinfo{author}{\bibfnamefont{I.}~\bibnamefont{Kim}}, \bibnamefont{and}
  \bibinfo{author}{\bibfnamefont{J.}~\bibnamefont{Rho}},
  \bibinfo{journal}{Physical Chemistry Chemical Physics}
  \textbf{\bibinfo{volume}{22}}, \bibinfo{pages}{2337} (\bibinfo{year}{2020}).

\bibitem[{\citenamefont{Popova et~al.}(2018)\citenamefont{Popova, Isayev, and
  Tropsha}}]{popova2018deep}
\bibinfo{author}{\bibfnamefont{M.}~\bibnamefont{Popova}},
  \bibinfo{author}{\bibfnamefont{O.}~\bibnamefont{Isayev}}, \bibnamefont{and}
  \bibinfo{author}{\bibfnamefont{A.}~\bibnamefont{Tropsha}},
  \bibinfo{journal}{Science advances} \textbf{\bibinfo{volume}{4}},
  \bibinfo{pages}{eaap7885} (\bibinfo{year}{2018}).

\bibitem[{\citenamefont{De~Yoreo et~al.}(2015)\citenamefont{De~Yoreo, Gilbert,
  Sommerdijk, Penn, Whitelam, Joester, Zhang, Rimer, Navrotsky, Banfield
  et~al.}}]{de2015crystallization}
\bibinfo{author}{\bibfnamefont{J.~J.} \bibnamefont{De~Yoreo}},
  \bibinfo{author}{\bibfnamefont{P.~U.} \bibnamefont{Gilbert}},
  \bibinfo{author}{\bibfnamefont{N.~A.} \bibnamefont{Sommerdijk}},
  \bibinfo{author}{\bibfnamefont{R.~L.} \bibnamefont{Penn}},
  \bibinfo{author}{\bibfnamefont{S.}~\bibnamefont{Whitelam}},
  \bibinfo{author}{\bibfnamefont{D.}~\bibnamefont{Joester}},
  \bibinfo{author}{\bibfnamefont{H.}~\bibnamefont{Zhang}},
  \bibinfo{author}{\bibfnamefont{J.~D.} \bibnamefont{Rimer}},
  \bibinfo{author}{\bibfnamefont{A.}~\bibnamefont{Navrotsky}},
  \bibinfo{author}{\bibfnamefont{J.~F.} \bibnamefont{Banfield}},
  \bibnamefont{et~al.}, \bibinfo{journal}{Science}
  \textbf{\bibinfo{volume}{349}}, \bibinfo{pages}{aaa6760}
  (\bibinfo{year}{2015}).

\bibitem[{\citenamefont{Hagan and Chandler}(2006)}]{hagan2006dynamic}
\bibinfo{author}{\bibfnamefont{M.~F.} \bibnamefont{Hagan}} \bibnamefont{and}
  \bibinfo{author}{\bibfnamefont{D.}~\bibnamefont{Chandler}},
  \bibinfo{journal}{Biophysical Journal} \textbf{\bibinfo{volume}{91}},
  \bibinfo{pages}{42} (\bibinfo{year}{2006}).

\bibitem[{\citenamefont{Wilber et~al.}(2007)\citenamefont{Wilber, Doye, Louis,
  Noya, Miller, and Wong}}]{wilber2007reversible}
\bibinfo{author}{\bibfnamefont{A.~W.} \bibnamefont{Wilber}},
  \bibinfo{author}{\bibfnamefont{J.~P.} \bibnamefont{Doye}},
  \bibinfo{author}{\bibfnamefont{A.~A.} \bibnamefont{Louis}},
  \bibinfo{author}{\bibfnamefont{E.~G.} \bibnamefont{Noya}},
  \bibinfo{author}{\bibfnamefont{M.~A.} \bibnamefont{Miller}},
  \bibnamefont{and} \bibinfo{author}{\bibfnamefont{P.}~\bibnamefont{Wong}},
  \bibinfo{journal}{The Journal of Chemical Physics}
  \textbf{\bibinfo{volume}{127}}, \bibinfo{pages}{085106}
  (\bibinfo{year}{2007}).

\bibitem[{\citenamefont{Whitelam and Jack}(2015)}]{whitelam2015statistical}
\bibinfo{author}{\bibfnamefont{S.}~\bibnamefont{Whitelam}} \bibnamefont{and}
  \bibinfo{author}{\bibfnamefont{R.~L.} \bibnamefont{Jack}},
  \bibinfo{journal}{Annual review of physical chemistry}
  \textbf{\bibinfo{volume}{66}}, \bibinfo{pages}{143} (\bibinfo{year}{2015}).

\bibitem[{\citenamefont{Gillespie}(2007)}]{gillespie2007stochastic}
\bibinfo{author}{\bibfnamefont{D.~T.} \bibnamefont{Gillespie}},
  \bibinfo{journal}{Annu. Rev. Phys. Chem.} \textbf{\bibinfo{volume}{58}},
  \bibinfo{pages}{35} (\bibinfo{year}{2007}).

\bibitem[{\citenamefont{McGrath et~al.}(2017)\citenamefont{McGrath, Jones, ten
  Wolde, and Ouldridge}}]{mcgrath2017biochemical}
\bibinfo{author}{\bibfnamefont{T.}~\bibnamefont{McGrath}},
  \bibinfo{author}{\bibfnamefont{N.~S.} \bibnamefont{Jones}},
  \bibinfo{author}{\bibfnamefont{P.~R.} \bibnamefont{ten Wolde}},
  \bibnamefont{and} \bibinfo{author}{\bibfnamefont{T.~E.}
  \bibnamefont{Ouldridge}}, \bibinfo{journal}{Physical Review Letters}
  \textbf{\bibinfo{volume}{118}}, \bibinfo{pages}{028101}
  (\bibinfo{year}{2017}).

\bibitem[{\citenamefont{Seifert}(2012)}]{seifert2012stochastic}
\bibinfo{author}{\bibfnamefont{U.}~\bibnamefont{Seifert}},
  \bibinfo{journal}{Reports on progress in Physics}
  \textbf{\bibinfo{volume}{75}}, \bibinfo{pages}{126001}
  (\bibinfo{year}{2012}).

\bibitem[{\citenamefont{Brown and Sivak}(2017)}]{brown2017allocating}
\bibinfo{author}{\bibfnamefont{A.~I.} \bibnamefont{Brown}} \bibnamefont{and}
  \bibinfo{author}{\bibfnamefont{D.~A.} \bibnamefont{Sivak}},
  \bibinfo{journal}{Proceedings of the National Academy of Sciences}
  \textbf{\bibinfo{volume}{114}}, \bibinfo{pages}{11057}
  (\bibinfo{year}{2017}).

\bibitem[{\citenamefont{Seifert}(2005)}]{seifert2005entropy}
\bibinfo{author}{\bibfnamefont{U.}~\bibnamefont{Seifert}},
  \bibinfo{journal}{Physical Review Letters} \textbf{\bibinfo{volume}{95}},
  \bibinfo{pages}{040602} (\bibinfo{year}{2005}).

\bibitem[{\citenamefont{Lecomte et~al.}(2007)\citenamefont{Lecomte,
  Appert-Rolland, and Van~Wijland}}]{lecomte2007thermodynamic}
\bibinfo{author}{\bibfnamefont{V.}~\bibnamefont{Lecomte}},
  \bibinfo{author}{\bibfnamefont{C.}~\bibnamefont{Appert-Rolland}},
  \bibnamefont{and}
  \bibinfo{author}{\bibfnamefont{F.}~\bibnamefont{Van~Wijland}},
  \bibinfo{journal}{Journal of statistical physics}
  \textbf{\bibinfo{volume}{127}}, \bibinfo{pages}{51} (\bibinfo{year}{2007}).

\bibitem[{\citenamefont{Ritort}(2008)}]{ritort2008nonequilibrium}
\bibinfo{author}{\bibfnamefont{F.}~\bibnamefont{Ritort}},
  \bibinfo{journal}{Advances in Chemical Physics}
  \textbf{\bibinfo{volume}{137}}, \bibinfo{pages}{31} (\bibinfo{year}{2008}).

\bibitem[{\citenamefont{Garrahan et~al.}(2009)\citenamefont{Garrahan, Jack,
  Lecomte, Pitard, van Duijvendijk, and van Wijland}}]{garrahan2009first}
\bibinfo{author}{\bibfnamefont{J.~P.} \bibnamefont{Garrahan}},
  \bibinfo{author}{\bibfnamefont{R.~L.} \bibnamefont{Jack}},
  \bibinfo{author}{\bibfnamefont{V.}~\bibnamefont{Lecomte}},
  \bibinfo{author}{\bibfnamefont{E.}~\bibnamefont{Pitard}},
  \bibinfo{author}{\bibfnamefont{K.}~\bibnamefont{van Duijvendijk}},
  \bibnamefont{and} \bibinfo{author}{\bibfnamefont{F.}~\bibnamefont{van
  Wijland}}, \bibinfo{journal}{Journal of Physics A: Mathematical and
  Theoretical} \textbf{\bibinfo{volume}{42}}, \bibinfo{pages}{075007}
  (\bibinfo{year}{2009}).

\bibitem[{\citenamefont{Speck et~al.}(2012)\citenamefont{Speck, Engel, and
  Seifert}}]{speck2012large}
\bibinfo{author}{\bibfnamefont{T.}~\bibnamefont{Speck}},
  \bibinfo{author}{\bibfnamefont{A.}~\bibnamefont{Engel}}, \bibnamefont{and}
  \bibinfo{author}{\bibfnamefont{U.}~\bibnamefont{Seifert}},
  \bibinfo{journal}{Journal of Statistical Mechanics: Theory and Experiment}
  \textbf{\bibinfo{volume}{2012}}, \bibinfo{pages}{P12001}
  (\bibinfo{year}{2012}).

\bibitem[{\citenamefont{Lecomte et~al.}(2010)\citenamefont{Lecomte, Imparato,
  and Wijland}}]{lecomte2010current}
\bibinfo{author}{\bibfnamefont{V.}~\bibnamefont{Lecomte}},
  \bibinfo{author}{\bibfnamefont{A.}~\bibnamefont{Imparato}}, \bibnamefont{and}
  \bibinfo{author}{\bibfnamefont{F.~v.} \bibnamefont{Wijland}},
  \bibinfo{journal}{Progress of Theoretical Physics Supplement}
  \textbf{\bibinfo{volume}{184}}, \bibinfo{pages}{276} (\bibinfo{year}{2010}).

\bibitem[{\citenamefont{Harris}(2015)}]{harris2015fluctuations}
\bibinfo{author}{\bibfnamefont{R.~J.} \bibnamefont{Harris}},
  \bibinfo{journal}{Journal of Statistical Mechanics: Theory and Experiment}
  \textbf{\bibinfo{volume}{2015}}, \bibinfo{pages}{P07021}
  (\bibinfo{year}{2015}).

\bibitem[{\citenamefont{Callen}(1985)}]{Callen_1985}
\bibinfo{author}{\bibfnamefont{H.~B.} \bibnamefont{Callen}},
  \emph{\bibinfo{title}{Thermodynamics and an introduction to
  thermostatistics}} (\bibinfo{publisher}{John Wiley \& Sons},
  \bibinfo{address}{New York}, \bibinfo{year}{1985}), \bibinfo{edition}{2nd}
  ed.

\bibitem[{\citenamefont{Carnot}(1824)}]{Carnot}
\bibinfo{author}{\bibfnamefont{S.}~\bibnamefont{Carnot}},
  \bibinfo{journal}{Reflections on the motive power of fire by Sadi Carnot and
  other papers on the Second Law of Thermodynamics by E. Clapeyron and R}
  (\bibinfo{year}{1824}).

\bibitem[{\citenamefont{Silberberg}(2007)}]{Ideal}
\bibinfo{author}{\bibfnamefont{M.~S.} \bibnamefont{Silberberg}},
  \emph{\bibinfo{title}{Principles of general chemistry}}
  (\bibinfo{publisher}{McGraw-Hill Higher Education New York},
  \bibinfo{year}{2007}).

\bibitem[{\citenamefont{Sutton et~al.}(1998)\citenamefont{Sutton, Barto, Bach
  et~al.}}]{RL}
\bibinfo{author}{\bibfnamefont{R.~S.} \bibnamefont{Sutton}},
  \bibinfo{author}{\bibfnamefont{A.~G.} \bibnamefont{Barto}},
  \bibinfo{author}{\bibfnamefont{F.}~\bibnamefont{Bach}}, \bibnamefont{et~al.},
  \emph{\bibinfo{title}{Reinforcement learning: An introduction}}
  (\bibinfo{publisher}{MIT press}, \bibinfo{year}{1998}).

\bibitem[{\citenamefont{Finkelstein}(1829)}]{Stirling}
\bibinfo{author}{\bibfnamefont{T.}~\bibnamefont{Finkelstein}},
  \bibinfo{type}{Tech. Rep.}, \bibinfo{institution}{AIAA-94-3951-CP}
  (\bibinfo{year}{1829}).

\bibitem[{\citenamefont{Mozurkewich and Berry}(1982)}]{Otto}
\bibinfo{author}{\bibfnamefont{M.}~\bibnamefont{Mozurkewich}} \bibnamefont{and}
  \bibinfo{author}{\bibfnamefont{R.~S.} \bibnamefont{Berry}},
  \bibinfo{journal}{Journal of Applied Physics} \textbf{\bibinfo{volume}{53}},
  \bibinfo{pages}{34} (\bibinfo{year}{1982}).

\bibitem[{\citenamefont{B{\o}hn et~al.}(2019)\citenamefont{B{\o}hn, Coates,
  Moe, and Johansen}}]{bohn2019deep}
\bibinfo{author}{\bibfnamefont{E.}~\bibnamefont{B{\o}hn}},
  \bibinfo{author}{\bibfnamefont{E.~M.} \bibnamefont{Coates}},
  \bibinfo{author}{\bibfnamefont{S.}~\bibnamefont{Moe}}, \bibnamefont{and}
  \bibinfo{author}{\bibfnamefont{T.~A.} \bibnamefont{Johansen}}, in
  \emph{\bibinfo{booktitle}{2019 International Conference on Unmanned Aircraft
  Systems (ICUAS)}} (\bibinfo{organization}{IEEE}, \bibinfo{year}{2019}), pp.
  \bibinfo{pages}{523--533}.

\bibitem[{\citenamefont{Melo and M{\'a}ximo}(2019)}]{melo2019learning}
\bibinfo{author}{\bibfnamefont{L.~C.} \bibnamefont{Melo}} \bibnamefont{and}
  \bibinfo{author}{\bibfnamefont{M.~R. O.~A.} \bibnamefont{M{\'a}ximo}}, in
  \emph{\bibinfo{booktitle}{2019 Latin american robotics symposium (LARS), 2019
  Brazilian symposium on robotics (SBR) and 2019 workshop on robotics in
  education (WRE)}} (\bibinfo{organization}{IEEE}, \bibinfo{year}{2019}), pp.
  \bibinfo{pages}{37--42}.

\bibitem[{\citenamefont{Wei et~al.}(2019)\citenamefont{Wei, Liu, Mashayekhy,
  and Decker}}]{wei2019mixed}
\bibinfo{author}{\bibfnamefont{H.}~\bibnamefont{Wei}},
  \bibinfo{author}{\bibfnamefont{X.}~\bibnamefont{Liu}},
  \bibinfo{author}{\bibfnamefont{L.}~\bibnamefont{Mashayekhy}},
  \bibnamefont{and} \bibinfo{author}{\bibfnamefont{K.}~\bibnamefont{Decker}},
  in \emph{\bibinfo{booktitle}{2019 IEEE Vehicular Networking Conference
  (VNC)}} (\bibinfo{organization}{IEEE}, \bibinfo{year}{2019}), pp.
  \bibinfo{pages}{1--8}.

\bibitem[{\citenamefont{Zhang et~al.}(2021)\citenamefont{Zhang, Zhang, Zhao,
  and Zou}}]{zhang2021image}
\bibinfo{author}{\bibfnamefont{L.}~\bibnamefont{Zhang}},
  \bibinfo{author}{\bibfnamefont{Y.}~\bibnamefont{Zhang}},
  \bibinfo{author}{\bibfnamefont{X.}~\bibnamefont{Zhao}}, \bibnamefont{and}
  \bibinfo{author}{\bibfnamefont{Z.}~\bibnamefont{Zou}},
  \bibinfo{journal}{Image and Vision Computing} \textbf{\bibinfo{volume}{108}},
  \bibinfo{pages}{104126} (\bibinfo{year}{2021}).

\bibitem[{\citenamefont{De~Vree and Carloni}(2021)}]{de2021deep}
\bibinfo{author}{\bibfnamefont{L.}~\bibnamefont{De~Vree}} \bibnamefont{and}
  \bibinfo{author}{\bibfnamefont{R.}~\bibnamefont{Carloni}},
  \bibinfo{journal}{IEEE Transactions on Neural Systems and Rehabilitation
  Engineering} \textbf{\bibinfo{volume}{29}}, \bibinfo{pages}{607}
  (\bibinfo{year}{2021}).

\bibitem[{\citenamefont{Schulman et~al.}(2015)\citenamefont{Schulman, Levine,
  Abbeel, Jordan, and Moritz}}]{TRPO}
\bibinfo{author}{\bibfnamefont{J.}~\bibnamefont{Schulman}},
  \bibinfo{author}{\bibfnamefont{S.}~\bibnamefont{Levine}},
  \bibinfo{author}{\bibfnamefont{P.}~\bibnamefont{Abbeel}},
  \bibinfo{author}{\bibfnamefont{M.}~\bibnamefont{Jordan}}, \bibnamefont{and}
  \bibinfo{author}{\bibfnamefont{P.}~\bibnamefont{Moritz}}, in
  \emph{\bibinfo{booktitle}{International conference on machine learning}}
  (\bibinfo{year}{2015}), pp. \bibinfo{pages}{1889--1897}.

\end{thebibliography}
\bibliographystyle{apsrev}

\end{document}